\documentclass[10pt,twocolumn,letterpaper]{article}

\usepackage[pagenumbers]{cvpr}

\usepackage{graphicx}
\usepackage{amsmath}
\usepackage{amssymb}
\usepackage{booktabs}
\usepackage{enumitem}

\usepackage[accsupp]{axessibility}  

\usepackage[pagebackref,breaklinks,colorlinks]{hyperref}

\usepackage[capitalize]{cleveref}
\crefname{section}{Sec.}{Secs.}
\Crefname{section}{Section}{Sections}
\Crefname{table}{Table}{Tables}
\crefname{table}{Tab.}{Tabs.}

\usepackage{pifont}
\newcommand{\cmark}{\ding{51}}%

\usepackage[table,xcdraw]{xcolor}
\definecolor{newcolor}{rgb}{.8,.349,.1}

\newcommand{\ours}{\mbox{OO-dMVMT}\xspace}

\newcommand{\shrecventidue}{\mbox{SHREC'22}\xspace}
\newcommand{\shrecdiciannove}{\mbox{SHREC'19}\xspace}

\newcommand\blfootnote[1]{%
  \begingroup
  \renewcommand\thefootnote{}\footnote{#1}%
  \addtocounter{footnote}{-1}%
  \endgroup
}

\begin{document}

\title{\ours: A Deep Multi-view Multi-task Classification Framework for Real-time 3D Hand Gesture Classification and Segmentation}

\author{
\begin{tabular}{ccc}
Federico Cunico\textsuperscript{*}&
Federico Girella\textsuperscript{*}&
Andrea Avogaro\textsuperscript{*}
\\
Marco Emporio\textsuperscript{*}&
Andrea Giachetti&
Marco Cristani \vspace{0.2cm}
\end{tabular}
\\
University of Verona\\
\tt\small{name.surname@univr.it}
}

\maketitle


\begin{abstract}
Continuous mid-air hand gesture recognition based on captured hand pose streams is fundamental for human-computer interaction, particularly in AR / VR. However, many of the methods proposed to recognize heterogeneous hand gestures are tested only on the classification task, and the real-time low-latency gesture segmentation in a continuous stream is not well addressed in the literature. For this task, we propose the On-Off deep Multi-View Multi-Task paradigm (\ours).
The idea is to exploit multiple time-local views related to hand pose and movement to generate rich gesture descriptions, along with using heterogeneous tasks to achieve high accuracy.
\ours extends the classical MVMT paradigm, where all of the  multiple tasks have to be active at each time, 
by allowing specific tasks to switch on/off depending on whether they can apply to the input.
We show that \ours defines the new SotA on continuous/online 3D skeleton-based gesture recognition
in terms of gesture classification accuracy, segmentation accuracy, false positives, and decision latency while maintaining real-time operation.
\vspace{-1.2em}
\end{abstract}

\blfootnote{\textsuperscript{*}The authors contributed equally to this paper}
\blfootnote{Code: \href{https://github.com/intelligolabs/OO-dMVMT}{https://github.com/intelligolabs/OO-dMVMT}}

\section{Introduction}
\label{sec:introiccv}

\begin{figure}
    \centering
    \includegraphics[width=\linewidth]{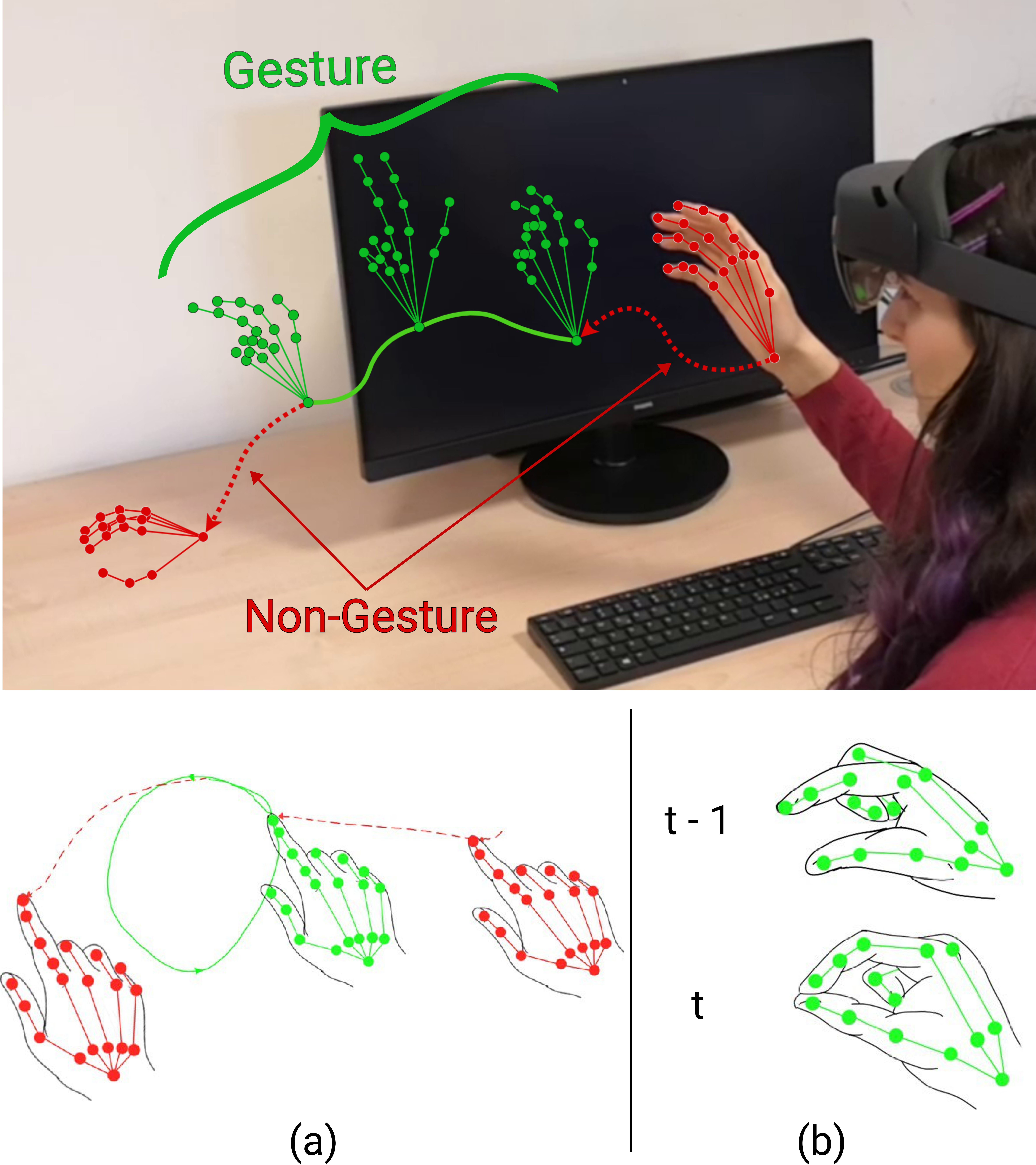}
    \caption{On top, the 3D hand gesture real-time classification and segmentation scenario: in red are non-gestures, in green the segmented gesture. The task is difficult: below, we show samples from \shrecdiciannove (a), where it is possible to see long gestures with a static pose of the hand and the correct start of the gesture  cannot be easily detected; in (b) a sample form \shrecventidue, where gestures express more pose dynamics.
    }
    \label{fig:teaser}
    \vspace{-1.2em}
\end{figure}

The current generation of Mixed Reality (MR) headsets (e.g. Meta Quest 2, Hololens 2, etc.) features accurate hand tracking capabilities, capturing finger poses at
high frame rates, and cheap sensors/APIs are available to enhance different applications with this data.
Real-time, reliable recognition of mid-air hand gestures is a fundamental ingredient for building 3D ``natural'' interfaces for MR, Human-Robot, and Human-Machine interaction in industry, home appliance control, and more. 

In these contexts, the gestures have to be detected and classified within a continuous stream of hand movements that can be captured by cameras or other sensors. 
A variety of skeleton-based architectures for hand gesture recognition have been proposed over the years, leveraging different methodologies such as multi-view learning~\cite{devineau2018deep,yang2019make,trivedi2022psumnet}, attentive models~\cite{maghoumi2018deepgru}, graph convolutional networks~\cite{hou2018spatial,li2019spatial,chen2019construct,liu2020disentangling,li2021two,shi2020decoupled,sampieri2022pose,cai2019exploiting}, LSTMs~\cite{avola2018exploiting}.
Most of these methods use features extracted from the data, proving the importance of enriching the input, rather than using all the raw data.

However, none of the methods proposed in the literature has the accuracy and a false positives' score good enough to guarantee a natural and reliable MR interaction. 
A missed or wrong gesture classification can potentially be dangerous for the users if they have to interact with the machinery of an industry, medical operation, or any delicate situation.

In order to build an effective model, we tackled the online gesture recognition problem using a powerful yet unexplored framework called Multi-View Multi-Task (MVMT) learning paradigm~\cite{he2011graphbased,lu2017multilinear}. 
In MVMT, complementary views of the object of interest serve to learn a rich feature embedding, that is fed into multiple tasks to borrow information across tasks that can potentially increase the predictive power of the learned models.
Multi-View (MV) learning~\cite{zhao2017multi} 
is a well-established direction in machine learning which
associates multiple feature sets (called ``views'') to the input data \eg{} given an image, extract spatial features, color features and frequency features to form a single data point.
Each data point is thus associated with $V$ sets of different features where each set corresponds to a view, allowing supervised or semi-supervised learning tasks to access a more complete data descriptor.
MV learning does not have to be confused with the multi-camera approach in which multiple camera views create their data point, that are then usually merged (synchronized) into a single one. 
Multi-Task (MT) learning aims to leverage useful information contained in multiple related tasks to help improve the generalization performance of at least one of the tasks~\cite{zhang2021survey}.

MVMT learning is a less explored field, which extends MV learning to the MT learning setting with $T$ tasks, where each task is a multi-view learning problem~\cite{he2011graphbased}. Most of the existing MVMT methods focus on proposing linear models for fitting the specific application requirements even though they are not suitable for common large-scale real-world problems in real environments. Only recently, non-linear deep models have been taken into account~\cite{wu2018dmtmv}.
In our case, we take into account multiple views related to hand geometry and its global movement to create a multi-view description. Furthermore, we raise the accuracy by exploiting the relatedness of multiple tasks such as gesture classification at different grains and regression of its start-end instants.

The issue arises when some tasks can't be performed under certain conditions, such as the regression of the start frame of a gesture when no gestures are active.
This invalidates the classical 
MVMT
framework, which requires all the tasks to be active at each time.
For this reason, we propose the On-Off deep Multi-View Multi-Task paradigm (\ours), which extends MVMT by allowing specific tasks to switch on/off depending on whether they can process the observation sample.

\ours deals exactly with this situation, introducing a mechanism that, at training time, switches tasks on and off depending on whether they can act on a specific data observation. 
Our contributions are the following:
\begin{itemize}[topsep=0.2em]
    \setlength\itemsep{-0.1em}
    \item We individuate MVMT as the correct framework to deal with real-time hand gesture recognition. 
    \item We extend MVMT to adapt to cases when some of the tasks cannot operate on a specific data sample, introducing \ours
    \item With \ours, we define the new state-of-the-art of hand gestures real-time classification and segmentation.
\end{itemize}

We show the performance of our method in the realistic scenario proposed by specific benchmarks 
(\shrecdiciannove~\cite{caputo2019shrec}, \shrecventidue~\cite{emporio2022shrec}), where gestures live within long sequences of non-significant hand movements from which they need to be distinguished.

\section{Related works}
\label{sec:relatediccv}

Hand gestures designed for AR/VR interaction have short but variable lengths, varying action granularity, and their semantics may depend on completely different factors (the static pose, the global hand trajectory, the fingers' articulation). 
While many classifiers have been proposed to recognize gestures of this kind from hand poses' sequences, most of them were not tested in a realistic scenario, but only tested on offline benchmarks like DHG14/28~\cite{de2016skeleton} and SHREC'17~\cite{de2017shrec}. These benchmarks only evaluate the labeling accuracy of pre-segmented gestures obtained with classifiers trained on similar data.

\noindent\textbf{Offline classification methods.}
For this task different network architectures have been proposed. Devineau et al.~\cite{devineau2018deep} proposed a multi-channel convolutional neural network with two feature extraction modules and a residual branch per channel. 
Maghoumi et al.~\cite{maghoumi2018deepgru} proposed DeepGRU, based on a set of stacked gated recurrent units (GRU) and a global attention model. 
Hou et al.~\cite{hou2018spatial} employed an end-to-end Spatial-Temporal Attention Residual Temporal Convolutional Network (STA-Res-TCN). 
Spatial-temporal GCN has been also adapted for hand gesture recognition~\cite{li2019spatial}.
Chen et al.~\cite{chen2019construct} built a fully connected graph from the hand skeleton and learned node features and edges via a self-attention mechanism that performs in both spatial and temporal domains. 
Liu et al.~\cite{liu2020disentangling} extracted optimized features from skeleton data by using a disentangled multi-scale aggregation scheme. %
Li et al.~\cite{li2021two} proposed a two-stream network to address the variable temporal scales of the gesture classes. The first stream extracts short-term temporal information and hierarchical spatial information, and the second one extracts long-term temporal information.
Shi et al.~\cite{shi2020decoupled} modeled the spatial-temporal dependencies between joints without the requirement of knowing their positions or mutual connections by employing a Spatial-Temporal Attention Network.

A key observation looking at the literature on segmented gesture classification is that, while there is no evidence of particular advantages in the accuracy obtained with different network architectures (recurrent, graph-convolutional, 1D convolutional networks), it seems that better classification performances have been obtained by feeding the networks with pre-computed features instead of raw data, often organized in multiple sets with different semantics.
Avola et al.~\cite{avola2018exploiting} used a stack of LSTM units fed with angles at fingers' joints, intra-finger angles, fingertips', and hand center displacements.
In \cite{chen2019mfa} global and local motion features, along with the skeleton sequence, are fed into different branches of a Long-Short
Term Memory (LSTM) network to get the predicted class of input gesture.
Yang et al.~\cite{yang2019make} proposed DDNet, a simple network based on 1D convolutions, fed with multiple features (linearized joints' distance matrix, joints speeds) with a motion summarising module to reduce noise from non-relevant frames. 
Trivedi et al.~\cite{trivedi2022psumnet} used multiple features (joints, bones, joints' velocity, bones velocity) to feed a Spatio-Temporal Relational Module.
The use of the pre-computed features results in lighter and easier to train networks and the parallel encoding of different sets of features can be interpreted as an application \textbf{multi-view} learning paradigm. 
Note that multi-view here does not mean that multiple features are related to different physical viewpoints (even if this idea has been proposed as well, for example, in \cite{li2023mvhanet}).

\noindent\textbf{Continuous (online) classification.}
To build effective interfaces, we need to perform continuous or ``online'' recognition, which requires performing multiple tasks: localizing the gestures in the pose stream and labeling them by only using past information, providing results with a small delay, and avoiding detection not corresponding to actual gestures (false positives).
Specific benchmarks have only recently been proposed. The first is
\shrecdiciannove~\cite{caputo2019shrec}, that tests continuous recognition, but features only simple dynamic gestures characterized by global trajectories without hand articulation (see Sec.~\ref{sec:sh19}).
The benchmark we adopted to train and evaluate our system is \shrecventidue~\cite{emporio2022shrec} where the gesture dictionary is heterogeneous, including static and dynamic gestures similar to those proposed in DHG14/28 and SHREC'17 (see Sec.~\ref{sec:sh22}).
This benchmark 
improves the previous SHREC'21~\cite{caputo2021shrec}, proposed by the same authors 
and with a similar structure, but solving some issues that, according to the website, made it unreliable for comparative tests.
For these benchmarks, training data are long sequences of pose streams with annotated start frames, end frames, and labels of the gestures executed and the task is to detect the occurrences of gestures in test sequences using past information only. SHREC'22 data are captured with an Hololens 2 headset, \textit{meaning that a recognizer trained for the task could be directly integrated into an interactive demo} using the same device to capture hand poses. 

Two classes of approaches can be adopted to handle the online task~\cite{escalera2017challenges}: \emph{direct} and \emph{indirect} methods.
Direct methods employ specialized heuristics based on speed, energy, or curvature (or trained networks) to find gesture boundaries and then send the extracted candidates to a classification module. 
Although this approach is not expected to work well in the case of complex motions/gestures~\cite{taranta2021machete}, it has been proposed in different implementations.
In \shrecdiciannove~\cite{caputo2019shrec}, Seg.LSTM1 uses an LSTM together with a specialized segmentation network.
The ST-GCN method used in \cite{caputo2021shrec} uses an energy-based segmentation approach adding several ad-hoc rules.
In the 2ST-GCN method of \cite{emporio2022shrec}, an energy-based detection module is combined with a fine-grained classifier, providing gesture/non-gesture discrimination.
Indirect methods perform a continuous classification (simultaneous detection and labeling). This can be done using pre-trained classifiers working with fixed-size input sub-sequences and sliding windows schemes or by using recurrent networks like Long-Short Time Memory (LSTM) networks or Gated Recurrent Units (GRU). 
A simplified version of DeepGRU~\cite{maghoumi2018deepgru}, combined with a smart data augmentation method and adapted for online detection, performed best in the \shrecdiciannove contest \cite{caputo2019shrec}, but a similar solution did not perform well on complex heterogeneous gestures~\cite{caputo2021shrec}. 
The use of sliding windows is a simple and popular solution for the online task. Two main ideas have been proposed in this context: 
train the classifier on segmented gestures, handling the unknown input gesture length in the online testing \cite{emporio21stronger}, or train it on fixed-size sub-parts (solution adopted by two methods presented in \cite{emporio2022shrec}).
In \cite{emporio21stronger} a modified DDNet~\cite{yang2019make} is trained with resampled, segmented gestures and randomly sampled non-gesture windows. The online testing is performed with windows of variable width and the results are combined with a voting procedure. 
The TN-FSM method proposed in \shrecventidue~\cite{emporio2022shrec} trains a Transformer Network to classify windows of 10 frames on subsets of the training sequences and uses a Finite State Machine to create the online prediction.
Causal TCN, the method providing the best accuracy in \cite{emporio2022shrec}, trains a temporal convolutional network on 20-frames windows labeled with gesture classes or non-gestures depending on their intersections with the annotated ground truth of the training set. 
We decided to apply a similar protocol for our online training framework and we tested many of the models proposed for the offline setting in the continuous task, training them with fixed-size windows and adding a non-gesture class. Within this framework, we finally developed our novel On-Off deep Multi-View Multi-Task (\ours) method providing enhanced performances with respect to the SotA classifiers. 
We show that with \ours using a sliding window approach we are able to perform not only real-time gesture classification but also gesture segmentation with SotA performances.

\section{Our pipeline}\label{sec:MC_hand}
We present our proposed \ours by first detailing the views and the tasks we have  considered (Sec.~\ref{sec:views} and Sec.~\ref{sec:Tasks}, respectively). Then, we present how to train the model (Sec.~\ref{sec:training}) and how to infer from it for the real-time hand classification problem (Sec.~\ref{sec:inference}). To fix the notation,  $\xi_{t}$ is the observation window  that ends at frame $t$ of length $W$, covering the frame interval $[t-W +1,t]$.

\subsection{The views}\label{sec:views}
Every observation training window $\xi_{t}$ is associated to $V=3$ views:  
1) \emph{Geometric layout}. We borrow from~\cite{yang2019make} the flattened Joint Collection Distances (JCD) features, which are location-viewpoint invariant. JCD computes the Euclidean distances between a pair of collective joints, and is of size $\binom{J}{2}$ for each frame of the observation window, resulting in a tensor of $\binom{J}{2} \times W$. 
2) \emph{Short-term slow motion} $M_{slow}$. 
$M_{slow}$ computes the 1-frame linear velocity of every single joint for all the joints, resulting in a tensor of size $J \times W-1$. 
3) \emph{Short-term fast motion} $M_{fast}$. 
$M_{fast}$ is similar to the short-term slow motion, but the linear velocity is computed every other frame (skipping the ones in between), 
with a  $J \times \frac{W}{2}-1$ resulting tensor. In practice, $M_{slow}$ and $M_{fast}$ model the short-term global motion of the skeleton in terms of speed. 
Following the  
multi-view learning paradigm, the three views JCD, $M_{slow}$ and $M_{fast}$ are each embedded into a $\mathbb{R}^{\frac{W}{2}\times8}$ dimensional space by a set of independent encoders 
and concatenated in a multi-view latent pattern $g_t$ of $\mathbb{R}^{\frac{W}{2}\times24}$ dimensions.

\subsection{The tasks}\label{sec:Tasks}
$T=4$ tasks have been considered: 1) \emph{SDN classification}. We classify gestures into three main categories: Static gestures (S), Dynamic gestures (D), and Non-Gestures (N). This macro classification is typical in heterogeneous hand gesture modeling~\cite{emporio2022shrec,caputo2021shrec}, and every gesture can fall in one of these superclasses. Static gestures require the user to fix the hand in a predetermined pose, keeping it still for a while. Dynamic gestures require the user to move the hand centroid, following a specific trajectory. Non-gestures are all those natural movements of the hand that occur between gestures and are the most difficult class to capture.
2) \emph{Fine-grained classification of gestures}. 
This task implements the fine-grained classification of the window $\xi_{t}$ considering all the   
$L$ classes into play (including the non-gesture class). This is the only task that is activated during testing time.
3) and 4) \emph{Start/end gesture frame regression}. These two tasks perform the regression on the frame indices $t_{start}$, $t_{end} \in [0,W-1]$ of the start and end of a gesture, respectively. The idea is to identify the 
patterns of the start and/or end of a gesture,
thus aiding the classification tasks.

These last two regression tasks could be unable to function,
due to the presence of 
observation windows which are completely contained in gesture (or non-gesture) streams.
This configuration, dealing with multi-views data and with some tasks which cannot be associated with some (or all) of the views of the input data, is novel in the MVMT literature.

\subsection{Model training}\label{sec:training}

Let $\xi_{t}$ be an input data instance sampled at time $t$, and $k$ the index of the $k$-th task.
We define the aggregated \ours objective function to minimize as: 
\begin{equation}\label{eq:OO0-MVMTobj}
F(\xi_{t}) = \sum_{k=1}^{T} c(k,t) \cdot F^{(k)}(g_{t},w_g, w_k)
\end{equation}

$F^{(k)}$ is the deep objective function associated with task $k$ controlled by $c(k,t)$, a boolean selector which is 1 if task $k$ can work with input $\xi_{t}$, otherwise $\xi_t$ does not contribute to the weights update for the task $k$. The decision of whether the task can work with the input is done at train time and depends on the semantics of $\xi_t$ and the task itself.
The term $g_{t}$ represents the multi-view representation of $\xi_{t}$ (see Sec.~\ref{sec:views}), $w_g$ are the multi-view parameters which generate $g_{t}$ starting from $\xi_{t}$, and $w_k$ are the $k$-th task parameters that generate the output for the $k$-th head.
The deep objective functions $F^{(k)}$ associated with our tasks are MSE 
for the regression tasks ($k=3,4$) and cross-entropy 
for the classification tasks ($k=1,2$). 
All the objective functions are uniformly weighted.

The idea is to treat the \ours model as an instance of asynchronous optimization~\cite{baytas2016asynchronous}, where the central server begins to update the shared model $w_g$ after it receives a gradient computation from one task node, without waiting for the other task nodes to finish their computations. In our case, the switching off of a task $k$ corresponds to having a delay for that task, while the other active tasks contribute to the model update.

\subsection{Model inference}\label{sec:inference}

\begin{figure*}[t]
    \centering
    \includegraphics[width=1\linewidth]{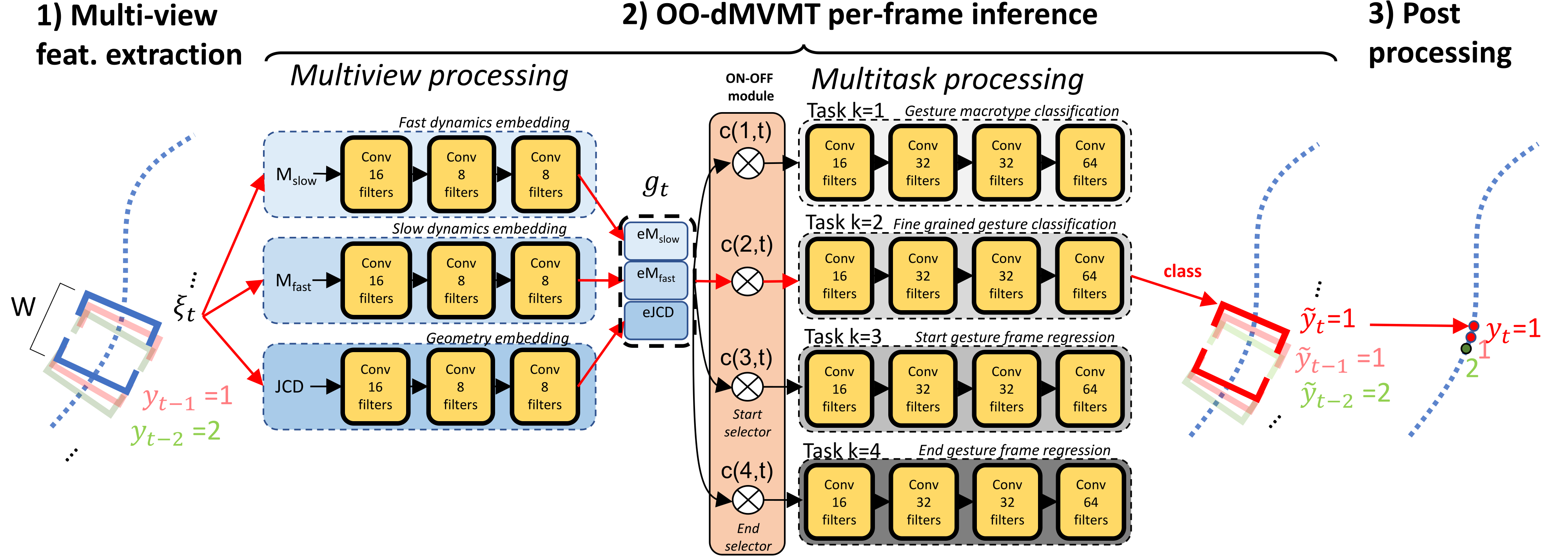}
    \caption{The pipeline of our framework, is composed of three steps: 1) Multi-view feature extraction, 2) \ours per-frame inference, 3) Post-processing. In red is highlighted the path used during the testing time, after the \ours model has been trained.}
    \label{fig:pipeline}
\end{figure*}

The pipeline of the framework \emph{at inference time} is summarized in Fig.~\ref{fig:pipeline} and consists of three steps:
in step 1) the observation window $\xi_{t}$ is sampled at frame $t$, and the multi-view description is extracted;  in step 2) $g_t$ flows into the fine-grained classification task, and a preliminary classification label $\tilde{y}_t$ is associated with the observation window $\xi_{t}$. The other three tasks are considered only during the training stage, as discussed in Sec.~\ref{sec:training}. 
3) a post-processing step is carried out, considering the previous $W-1$ 
preliminary classifications, 
in order to remove spurious ones. 
It is implemented by looking backward at $W$ frames, assigning the class label $y_t = l$ to the observation window $\xi_{t}$ if 
$l\in\{0,\ldots,L\}$
is the most frequent label in the last $W$ preliminary classification labels $\tilde{y}_t$.
Step 3 ensures a more stable output, as the single window classifications may be noisy if the window is small and the gesture dictionary includes complex examples with similar subparts.
It is worth noting that $W$ is an important parameter of the method that needs to be tuned based on the specific characteristics of the gestures' dictionary: it needs to be long enough to capture long gestures, but as short as possible to reduce input delay.

\section{Experiments}
The experiments are organized as follows: in Sec.~\ref{sec:sh22}, we detail the popular \shrecventidue~\cite{emporio2022shrec} and \shrecdiciannove~\cite{caputo2019shrec} datasets, together with the associated gesture classification metrics; in Sec.~\ref{sec:shrec22Res} we report the results on the \shrecventidue benchmark; in Sec.~\ref{sec:shrec19Res}  we focus on \shrecdiciannove; finally, we present multiple ablative studies in Sec.~\ref{sec:ablation}. 
As comparative approaches, we consider several graph-based architectures like
SeS-GCN~\cite{sampieri2022pose} and MS-G3D~\cite{liu2020disentangling}, the attention-based networks DG-STA~\cite{chen2019construct} and DSTA~\cite{shi2020decoupled}, a 1D convolutional network specifically designed for hand gestures, the Double-feature Double-motion Network (DDNet)~\cite{yang2019make}, and a lightweight action recognition network (PSUMNET~\cite{trivedi2022psumnet}).
These approaches add to the competitors which can be found in the \shrecventidue challenge~\cite{emporio2022shrec} and \shrecdiciannove challenge~\cite{caputo2019shrec}, for a total of 13 comparative methods.

\subsection{Datasets and evaluation metrics}
\paragraph{\shrecventidue.}
\label{sec:sh22}

The \shrecventidue benchmark~\cite{emporio2022shrec} features continuous recordings of 3D hand poses captured in simulated Mixed Reality interactions with an Hololens 2 device. The training set includes 144 sequences, with a total of 36 occurrences for each of the 16 gesture classes, interleaved with non-significant hand movements (non-gestures).
The testing set has the same cardinalities. Gestures belong to four categories: static, dynamic coarse, dynamic fine, and periodic. Each sequence 
is annotated with start frames, end frames, and gesture labels.
 Training and testing sequences have been captured by different subjects.
The benchmark evaluation protocol requires that the recognizer outputs sequences' annotations with the list of recognized gestures, their labels, and the predicted start and end frames. 
Since online prediction uses time samples after the estimated start, storing this delay is necessary for evaluating interface response time.
We follow the protocol and make use of the official evaluation code provided in the contest's repository~\cite{emporio2022shrec}.
The metrics are the following:

\noindent\textbf{Jaccard Index (JI)}: the average relative overlap between the ground truth and the predicted 
labels for the input sequences. It is used in many continuous classification tasks, but it does not evaluate 
the ability to avoid multiple activations for a single gesture or small noisy activations.

\noindent\textbf{Detection rate (DR)}: the ratio between the number of correctly detected gestures and the total number of gestures in the input sequences. A gesture is considered correctly detected if it has a temporal intersection (referred to as Minimum Overlap Ratio) with the ground truth greater than 50\% of the true interval, does not last more than twice the real duration, and has the same label. The gestures predicted by the recognizer but not corresponding to ground truth ones are defined as false positives.

\noindent\textbf{False positive score (FP)}: defined as the ratio between the number of false positives and the total number of gestures.

\noindent\textbf{Minimal detection delay (Delay)}: the delay metric is the difference in frames between the gesture start and the last frame used for the prediction.

\paragraph{\shrecdiciannove.}
\label{sec:sh19}
The \shrecdiciannove benchmark~\cite{caputo2019shrec} is focused on dynamic coarse gestures, characterized by the trajectory of the joints following specific 2D patterns (V-mark, X-mark, Caret, Square, Circle).
Hand trajectories have been captured with a Leap Motion sensor on users performing simulated interactive tasks in VR. The dataset is composed of 195 sequences, each one containing a single gesture 
surrounded by non-meaningful movements labeled as non-gestures. Each sequence
includes the 3D coordinates of the hand joints and the quaternions defining bones' orientation.  

The evaluation method provided in this benchmark considers an inference as correct if the correct predicted gesture is within 2.5 seconds from the ground truth gesture time window.
The metrics used for the evaluation of this benchmark are the DR, FP (defined as for \shrecventidue), and the inference time, intended as the time needed to perform a model inference for labeling a single frame. 

It is worth noting that \shrecventidue and \shrecdiciannove have similar 
evaluation
, but quite different dictionaries (heterogeneous types and duration of the classes in the first case, homogeneous in the second) and characteristics (acquisition devices, hand skeleton model, average gesture length).

\subsubsection{Implementation details}
The architecture of the network is shown in Fig.~\ref{fig:pipeline}. The view encoders are independent series of 1D Conv layers. 
Each task branch is a series of 1D Conv layers except for the last layers, which are fully-connected layers (with one final neuron for regression, three for SDN, and $L$ for the fine-grained classifier).
We adopted the Adafactor~\cite{shazeer2018adafactor} with an initial LR of 0.004 over $100$ epochs on 2 NVIDIA RTX3090.

\subsubsection{Real-time 3D hand pose}

As far as we know, the sequences in the benchmarks have been acquired with off-the-shelf devices and there are no guarantees on the accuracy of the tracking. 
This can possibly be one of the reasons for the non-optimal results of many methods, and \ours achieving better scores shows that it may be more robust against tracking errors. Improvements in hand tracking systems~\cite{han2020megatrack,han2022umetrack} could further reduce the errors without changing the algorithm.

\begin{table*}[t]
\caption{
Classification results of \shrecventidue benchmark. The 
metrics are the detection rate (DR), false positives scores (FP), the Jaccard Index (JI), the 
delay in frames, and per-frame processing time.
In brackets are the standard deviations.
Columns SV/MV refers to the single-view or multi-view learning approach, while ST/MT refers to single-task or multi-task learning approach.
}
\label{tab:SHREC22_results_f}
\centering
\begin{tabular}{lccccccccc}
\toprule 
\textbf{Method} & \textbf{DR} $\uparrow$ & \textbf{FP} $\downarrow$ & \textbf{JI} $\uparrow$ & \textbf{Delay(fr.)} $\downarrow$ & \textbf{time(ms)} $\downarrow$ & SV & MV & ST & MT \\
\midrule

DeepGRU~\cite{maghoumi2018deepgru} (2018) & 0.26 \textit{(.14)} & 0.25 \textit{(.23)} & 0.21 \textit{(.09)} & 8.00 & $3.1$ & \cmark & & \cmark &  \\
DG-STA~\cite{chen2019construct} (2019) & 0.51 \textit{(.10)} & 0.32 \textit{(.20)} & 0.40 \textit{(.20)} & 8.00 & 4.2  & \cmark & & \cmark & \\
SeS-GCN~\cite{sampieri2022pose} (2022) & 0.60 \textit{(.13)} & 0.16 \textit{(.09)} & 0.53 \textit{(.13)} & 8.00 & 1.8  & \cmark & & \cmark &  \\
PSUMNET~\cite{trivedi2022psumnet} (2022) & 0.62 \textit{(.14)} & 0.24 \textit{(.15)} & 0.52 \textit{(.15)} & 8.00 & 24.4 & & \cmark & \cmark &  \\
MS-G3D~\cite{liu2020disentangling} (2020) & 0.68 \textit{(.11)} & 0.21 \textit{(.15)} & 0.57 \textit{(.14)} & 8.00 & 29.3  & \cmark & & \cmark &  \\
Stronger~\cite{emporio2022shrec} (2022) & 0.72 \textit{(.11)} & 0.34 \textit{(.26)} & 0.59 \textit{(.18)} & 14.79 & 100.0  & & \cmark & \cmark &  \\
DSTA~\cite{shi2020decoupled} (2020) & 0.73 \textit{(.07)} & 0.24 \textit{(.13)} & 0.61 \textit{(.12)} & 8.00 & 9.2  & \cmark & & \cmark &  \\
2ST-GCN 5F~\cite{emporio2022shrec} (2022) & 0.74 \textit{(.12)} & 0.23 \textit{(.05)} & 0.61 \textit{(.11)} & 13.28 & 2.1  & \cmark & & & \cmark \\
TN-FSM+JD~\cite{emporio2022shrec} (2022) & 0.77 \textit{(.06)} & 0.23 \textit{(.12)} & 0.63 \textit{(.03)} & 10.00  & $4.6$  & & \cmark & \cmark &  \\
Causal TCN~\cite{emporio2022shrec} (2022) & 0.80 \textit{(.15)} & 0.29 \textit{(.22)} & 0.68 \textit{(.24)} & 19.00 & $28.0$  & \cmark & & \cmark &  \\
DDNet~\cite{yang2019make} (2019) & 0.88 \textit{(.06)} & 0.16 \textit{(.18)} & 0.78 \textit{(.14)} & 8.00 & 2.2  & & \cmark & \cmark &  \\

\midrule
FG + SDN & 0.86 \textit{(.06)} & 0.17 \textit{(.06)} & 0.75 \textit{(.11)} & 8.00 & 4.1  & & \cmark & & \cmark \\
FG + SDN + GC & 0.90 \textit{(.11)} & 0.10 \textit{(.11)} & 0.83 \textit{(.13)} & 8.00 & 4.0  & & \cmark & & \cmark \\
\textbf{\ours} & \textbf{0.92 \textit{(.06)}} & \textbf{0.09 \textit{(.09)}} & \textbf{0.85 \textit{(.11)}} & 8.00 & 4.1  & & \cmark & & \cmark \\
\bottomrule

\end{tabular}%
\end{table*}

\subsection{\shrecventidue results}\label{sec:shrec22Res}
In Tab.~\ref{tab:SHREC22_results_f} we report the official gesture classification metrics as provided by the evaluation code. For each comparative model, we look for its best parameters. The $W$ parameter should be as small as possible, to lower input delay, but large enough to capture significant information for longer gestures. In our case, we found $W=16$ to be an optimal choice for \shrecventidue.

\begin{figure}
    \centering
    \includegraphics[width=\linewidth]{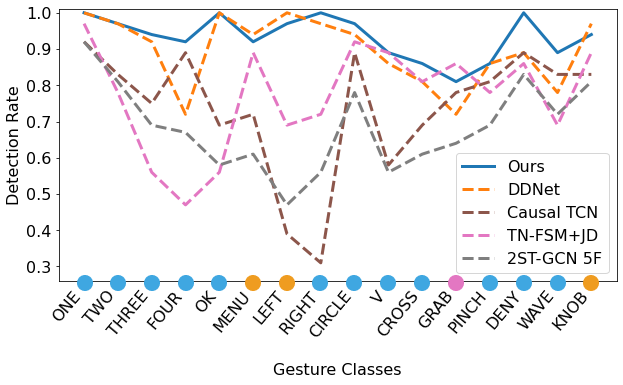}
    \caption{Detection rate per gesture class for \shrecventidue. Points are connected with a line for clarity. On the x-axis, a colored dot indicates, for each class, the method that achieved the best detection rate.
    Only the best 5 methods by DR are shown for clarity.}
    \label{fig:dr_gesture}
\end{figure}

\ours defines the new SOTA in all the metrics, with a fast inference time (4.1ms), and very low variances. 
Fig.~\ref{fig:dr_gesture} highlights the detection rate per gesture type of the five best performing models on \shrecventidue.
Notably, \ours is the most consistent top performer on this benchmark, with a detection rate that spans between 0.82 and 1, which is the thinnest min-max score range. 

\begin{figure}
    \centering
    \includegraphics[width=\linewidth]{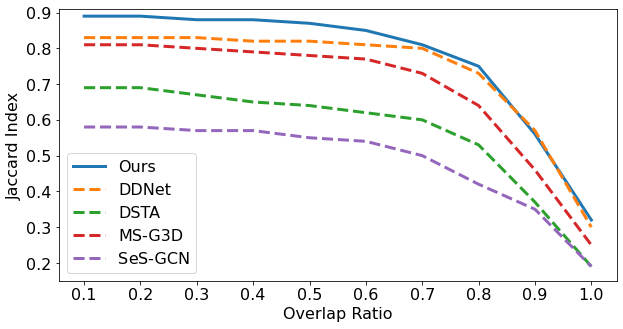}
    \caption{Jaccard index for \shrecventidue as a function of the minimum overlap ratio. For clarity, we show only the best 5 methods. 
    }
    \label{fig:ji_mor}
\end{figure}

An important parameter of the evaluation metrics for \shrecventidue~\cite{emporio2022shrec} is the \emph{minimum overlap ratio} (MOR) of the detected gestures with the ground truth. When $\text{MOR} = 1$ the predicted gesture needs to completely envelop the ground truth, while a value closer to $\text{MOR} = 0$ relaxes this constraint. Like in the original evaluation protocol, the results are evaluated with $\text{MOR}=0.5$.
Fig.~\ref{fig:ji_mor} shows the Jaccard Index (described in Sec.~\ref{sec:sh22}) as a function of the Minimum Overlap Ratio. Obviously, the higher the MOR, the lower the JI of all the approaches. Until $\text{MOR} = 0.7$, \ours clearly dominates the competitors. 

\begin{figure}
    \centering
    \includegraphics[width=\linewidth]{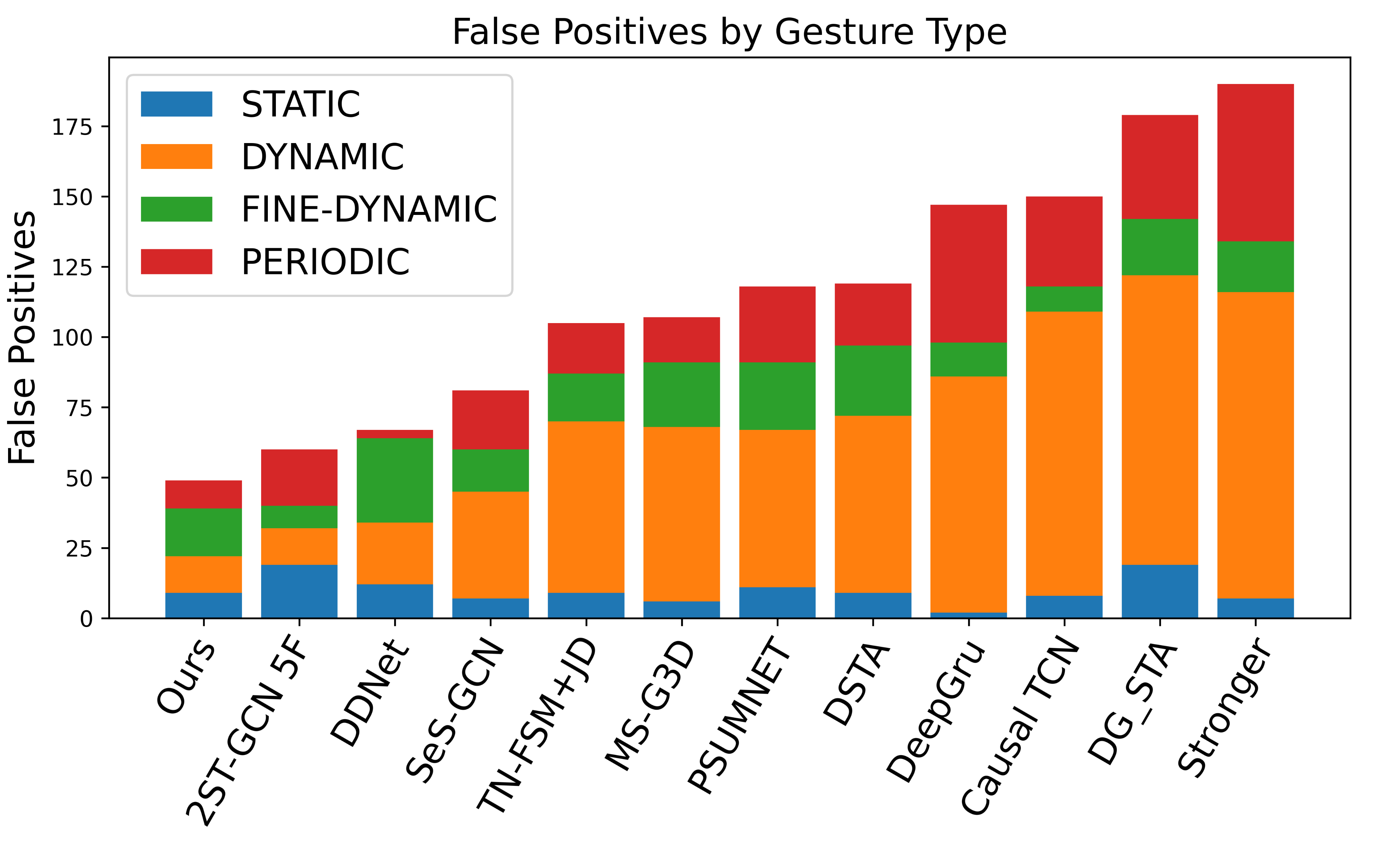}
    \caption{Stacked False Positives for \shrecventidue per model. The gestures are divided into 4 types (colors). The FPs for each type are then summed to show the total value. Lower is better.}
    \label{fig:fp_gesture_type}
\end{figure}

Of particular importance is the FP results, as they would trigger unwanted actions making the interaction frustrating. 
In Fig.~\ref{fig:fp_gesture_type}, a stacked barplot showing the total FPs for each model is presented (lower is better). Interestingly, the macro category which in general attracts the most false positives is the dynamic one, probably due to the fact that a movement of the limb is necessary to start this kind of gesture.
As soon as we move towards more effective approaches, it is easy to see that the number of false positive gestures considered as dynamic gestures diminishes drastically. Our approach has a balanced performance on all the gesture macro categories.

\subsection{\shrecdiciannove results}\label{sec:shrec19Res}
Results of our tests on the \shrecdiciannove benchmark~\cite{caputo2019shrec} are shown in Tab.~\ref{tab:SHREC19_results_f}. Some of the metrics are missing due to the unavailability in the original evaluations. 
The metrics differ slightly from the ones used in \shrecventidue~\cite{emporio2022shrec} due to the differences between \shrecdiciannove benchmark to its '22 counterpart, as discussed in Sec.~\ref{sec:sh19}.
For these tests, we chose a larger window size, $W = 40$. 
This is motivated by  
the longer average length of the gestures ($\sim 45$ frames) and by the fact that no short gestures are included in the dictionary.
The results are consistent with the ones presented in Sec.~\ref{sec:shrec22Res} and show our model achieving state-of-the-art performances on almost all of the metrics, except for the inference time.

\begin{table}[t]
\caption{
The metrics are the detection rate 
, false positives scores
, and inference time. 
Parameters 
with (*) comes 
from \cite{caputo2019shrec} and may not be accurate. 
Dash (--) values were not reported in the original benchmark. Standard deviation is reported between brackets. 
}
\label{tab:SHREC19_results_f}
\centering
\begin{tabular}{lccccc}
\toprule
\textbf{Method} & \textbf{DR} $\uparrow$ & \textbf{FP} $\downarrow$ & \textbf{time(ms)} $\downarrow$ \\
\midrule
PSUMNET~\cite{trivedi2022psumnet} & 0.64 \textit{(.21)} & 0.22 \textit{(.20)} & 25.0 \\
MS-G3D~\cite{liu2020disentangling} & 0.69 \textit{(.24)} & 0.25 \textit{(.22)} & 30.3 \\
SeS-GCN~\cite{sampieri2022pose} & 0.75 \textit{(.20)} & 0.12 \textit{(.13)} & 2.0 \\
SW 3-cent~\cite{3centrec} & 0.76 \textit{(--)} & 0.19 \textit{(--)} & 3.0* \\
DSTA~\cite{shi2020decoupled} & 0.81 \textit{(.11)} & 0.08 \textit{(.07)} & 8.8 \\
DG-STA~\cite{chen2019construct} & 0.81 \textit{(.11)} & 0.07 \textit{(.05)} & 4.2 \\
DDNet~\cite{yang2019make} & 0.82 \textit{(.13)} & 0.10 \textit{(.09)} & 2.2 \\
uDeepGRU~\cite{caputo2019shrec} & 0.85 \textit{(--)} & 0.10 \textit{(--)} & 3.0* \\
\midrule

\textbf{\ours} & \textbf{0.88 \textit{(.04)}} & \textbf{0.05 \textit{(.04)}} & 5.8 \\
\bottomrule
\end{tabular}%
\end{table}

\subsection{Ablation studies}\label{sec:ablation}
In this section, we show the impacts of our Multi-Task Learning On-Off paradigm, as well as discuss the impact of a correct training procedure for the regression heads in a Missing View scenario.

\subsubsection{Task head removal}
In this study, we explore the performance of \ours with some tasks removed. 
Results are in Tab.~\ref{tab:ablation}, where FG stands for fine-grained classification head only, FG + GS/GE adds the regression heads, and FG + SDN adds the coarse classification head. 
The FG model shows better performances than most methods of Tab.~\ref{tab:SHREC22_results_f}.
In the table we can see a pattern suggesting that, by introducing the multi-view approach, the performances tend to increase.
The SDN module slightly decreases the performances but ameliorates the standard deviations.
Surprisingly, the sole addition of the regression heads to the FG head drastically reduces the performances. 
This is probably due to the low task relatedness (start/end gestures are independent of the specific gesture classes), creating competing gradients during the model optimization. However, adding the SDN head allowed for reaching the optimal balance between classification and regression, with the latter improving the overall performance. It is worth noting that a grid search for the weights of the single tasks has been carried out, showing that uniform weighting was the best setup, and that each different task had a role in defining the optimal solution. 
Therefore, our joint \ours appears to be the right task ensemble. 
Additionally, we removed the regression heads in favor of a binary classifier (GC) trained to classify if a window contains the start or end of a gesture. 
As opposed to the always-active classification performed by GC, the On-Off regression procedure
produces better performance and more stable results.

\begin{table}[t]
\caption{Classification results on \shrecventidue with different Multi-task heads. Standard deviation is reported between brackets. 
}
\label{tab:ablation}
\centering
\begin{tabular}{lccc}
\toprule 
\textbf{Method} & \textbf{DR} $\uparrow$ & \textbf{FP} $\downarrow$ & \textbf{JI} $\uparrow$\\
\midrule
FG & 0.88 \textit{(.06)} & 0.16 \textit{(.18)} & 0.78 \textit{(.14)}\\
FG + GS/GE & 0.52 \textit{(.35)} & 0.48 \textit{(.31)} & 0.38 \textit{(.32)}\\
FG + SDN & 0.86 \textit{(.06)} & 0.17 \textit{(.06)} & 0.75 \textit{(.11)}\\
FG + SDN + GC & 0.90 \textit{(.11)} & 0.10 \textit{(.11)} & 0.83 \textit{(.13)}\\
\midrule
\textbf{\ours} & \textbf{0.92 \textit{(.06)}} & \textbf{0.09 \textit{(.09)}} & \textbf{0.85 \textit{(.11)}}\\
\bottomrule

\end{tabular}%
\end{table}

\subsubsection{The On-Off regression head}
Since the regression heads are the ones switched on and off by \ours, we further investigate them. In particular, we trained 
with different On-Off policies for the regression heads.
The Index Error simulates inaccurate gesture segmentation by randomly changing the gesture start/end index in window $\xi_t$ containing them.
The Window Error is more dramatic since it randomly switches on/off the heads with a probability $p=0.5$. It induces two kinds of errors since a wrongly activated window also implies a wrongly indexed gesture start.
In Tab.~\ref{tab:regress} the results confirm our expectations: 
both errors degrade the performance, with the Window Error having a worse impact compared to the Index Error.

\begin{table}[t]
\caption{Classification results on \shrecventidue with different Regression Heads training procedures. Standard deviation is reported between brackets. 
} 
\label{tab:regress}
\centering
\begin{tabular}{lccc}
\toprule 
\textbf{Method} & \textbf{DR} $\uparrow$ & \textbf{FP} $\downarrow$ & \textbf{JI} $\uparrow$\\
\midrule
Window Error & 0.76 \textit{(.33)} & 0.24 \textit{(.55)} & 0.65 \textit{(.31)}\\
Index Error & 0.83 \textit{(.14)} & 0.20 \textit{(.21)} & 0.72 \textit{(.19)}\\
\midrule
\textbf{\ours} & \textbf{0.92 \textit{(.06)}} & \textbf{0.09 \textit{(.09)}} & \textbf{0.85 \textit{(.11)}}\\
\bottomrule

\end{tabular}%
\end{table}

\section{Limitations}
Despite the results being SotA, the \ours framework presents a few limitations. Even though we achieve the best false positive performances as highlighted in Tab.~\ref{tab:SHREC22_results_f} we note that, for the fine-dynamic gestures, the false positives are higher than some of the comparative approaches. 
More research is needed to improve it with additional views or determine if hand acquisition noise is the issue.
Although our method has a longer execution time compared to other methods, it is still fast enough to be suitable for real-time applications.
Finally, the delay of our approach is heavily dependent on the value of $W$ as explained in Sec.~\ref{sec:inference}.

\section{Conclusions}
Our \ours paradigm for facing real-time classification and segmentation of complex hand gestures seems to be ideal: the geometric pose of a hand and its movement are views that are naturally complementary; 
at the same time,
the concurrent processing of multiple tasks is intuitive, mimicking how humans decode a gesture. For example, shaking hands requires that one understands 
both
the start of the gesture from the other partner (to initiate a proper reaction)
and the nature of the gesture. The SotA results on all the available benchmarks in the literature further promote our proposal. Our study, in addition, may help in the design of novel hand gestures: Fig.~\ref{fig:dr_gesture} indicates that, regardless of the specific approach, actions like ``grab'' give a local minimum in terms of detection rate. Future work will be devoted to forecasting the gesture, in order to 
delete the small delay (8 frames) that we are requiring here.

\section*{Acknowledgment}
This research is under the National Recovery and Resilience Plan (NRRP), Mission 4, Component 2 Investment 1.4, funded by the European Union – NextGeneration EU, project “Interconnected Nord-Est Innovation Ecosystem”.

{\small
\bibliographystyle{ieee_fullname}
\bibliography{main}
}

\end{document}